\documentclass[letterpaper]{article}
\usepackage[preprint]{aaai2027}
\usepackage[hyphens]{url}
\usepackage{graphicx}
\usepackage{natbib}
\usepackage{caption}
\usepackage{booktabs}

\title{TIP-Search: Time-Predictable Inference Scheduling for Market Prediction under Uncertain Load}
\author{
Xibai Wang
}
\affiliations{
NeuroQuant Labs Limited\\
Hong Kong SAR, China\\
xibai.wang@neuroquantlabs.com
}

\begin{document}
\raggedbottom
\maketitle

\begin{abstract}
Real-time market prediction services need correct predictions before a decision deadline; a correct prediction delivered late is not usable. TIP-Search studies time-predictable inference scheduling over fixed market predictors under uncertain load. It filters conformal latency-quantile feasible models, dispatches over finite workers, and uses shielded constrained online experts to trade accuracy, queue pressure, and deadline risk. On the optimized deployable pool, TIP-Search reaches 0.994 raw accuracy and 0.991 timely accuracy. On official TLOB FI-2010 h=10, TIP-Search++ raises timely accuracy from 0.156 to 0.239 and deadline satisfaction from 0.391 to 0.962. In matched h10 profiled systems replay, OCO-ACPO reaches 0.303 timely accuracy and 0.951 deadline satisfaction, with paired gains over RAMSIS/SneakPeek/utility-style comparators of $+0.00285$ timely accuracy ($p=0.0118$) and $+0.0146$ deadline satisfaction ($p=1.5{\times}10^{-5}$). SA-OCO-ACPO improves timely/deadline service by 0.188--0.417 over CPO under nonstationary stress. The claim is a systems scheduling result, not a broad LOB classifier leaderboard.
\end{abstract}

\section{Introduction}

Limit-order-book predictors such as DeepLOB \cite{zhang2019deeplob} improve market-state classification, but deployment also requires choosing a trained predictor that can finish before the remaining deadline. High-accuracy models can become unusable when queueing makes them late; fast models can degrade raw accuracy. TIP-Search studies this online scheduling tradeoff.

We use time-predictable to denote profile-audited inference selection that exposes latency feasibility and deadline risk under uncertain load. This is not a hard real-time worst-case guarantee, and the paper proposes a scheduler over fixed predictors, not a new LOB classifier or leaderboard.

\begin{figure}[!t]
\centering
\includegraphics[width=0.82\columnwidth]{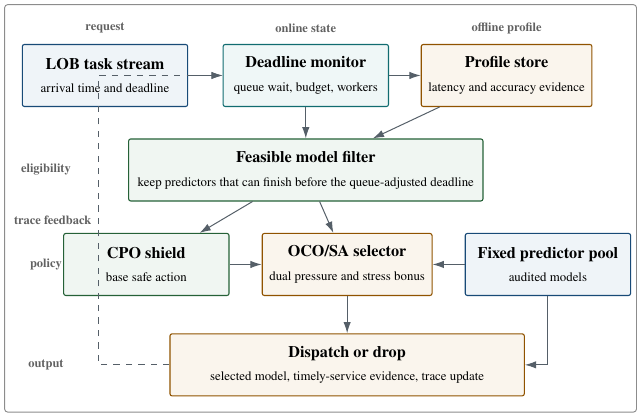}
\caption{TIP-Search dispatch architecture: deadline/queue-state filtering feeds the CPO shield and OCO/SA selector, which route a fixed predictor or drop the task while recording service evidence.}
\label{fig:architecture}
\end{figure}

TIP-Search is a time-predictable, deadline-aware inference scheduler: classifier accuracy is input evidence, while timely accuracy, $\Pr(\mathrm{correct}\wedge\mathrm{on\ time})$, is the router objective. The paper-facing official h10 controller, TIP-Search-OCO-ACPO, wraps TIP-Search-CPO with a base-action shield, finite CPO experts, and projected drop/response/debt dual pressure. SA-OCO-ACPO adds interval stress adaptation for nonstationary load. CPO, ACPO, OPT, LD, VQ, Q, and TIP-Search++ remain ablations that remove pressure, backlog, debt, or lower-confidence-bound components. Evaluation is scoped to executable local DeepLOB checkpoints for Binance BTC/USDT, FI-2010 \cite{ntakaris2018benchmark}, and LOBSTER-style \cite{huang2011lobster} inputs, plus source-verified official TLOB FI-2010 h=10 evidence.

The central thesis is that deployable prediction quality is a joint property of classifier correctness and service feasibility. Figure~\ref{fig:architecture} summarizes the service-level dispatch path; it is not a claim of live trading deployment. TIP-Search-OCO-ACPO and SA-OCO-ACPO are evaluated only as online routers over executable predictors and finite worker capacity.

The paper makes three bounded contributions:
\begin{itemize}
    \item A constrained online scheduler with finite-worker queue state, queue-adjusted budgets, latency-quantile feasibility, base CPO fallback, shielded finite experts, projected drop/response/debt dual traces, and an SA extension for interval-stress adaptation.
    \item A service-evaluation suite that reports raw accuracy as a classifier diagnostic and timely accuracy, deadline satisfaction, drop rate, queue wait, response time, and deadline debt as deployment metrics under local, official h=10, and nonstationary stress settings.
    \item A conservative evidence boundary separating scoped official h=10 classifier sanity checks, router timely-service evidence, queue-policy ablations, and unsupported external classifier targets.
\end{itemize}

The paper-facing boundary is narrow by design: dispatch and OCO-ACPO/SA-OCO-ACPO/ACPO/CPO claims are systems claims supported by local reruns, queue-policy evidence, official h10 timely-service gates, profiled systems replay, stress evidence, and load/trace/live-worker checks. LiT, ViT-LOB, TransLOB, LOBCAST, RAMSIS, SneakPeek, unsupported TLOB horizons, and cross-source TLOB remain excluded from classifier-SOTA claims unless verified official artifacts and horizon-specific gates become available.

\section{Problem Setting}

We consider a stream of inference tasks served by a finite-worker inference pool. Each task has an arrival time, a source label when available, and a deadline budget. The scheduler receives a finite model pool and the current worker availability vector. For each model, the local evidence package stores a profiled latency and an accuracy estimate derived from evaluation summaries. At dispatch, the service assigns the task to the earliest available worker, observes the resulting queue wait, computes the queue-adjusted budget as the remaining budget after queue wait, and must either select one model or drop the task when no model is admissible.

This setting differs from classical hard real-time scheduling \cite{liu1973scheduling}. The model pool is heterogeneous, the service time is profiled rather than analytically fixed, and prediction quality matters in addition to deadline feasibility. Therefore, the paper does not claim a hard real-time guarantee. Deadline satisfaction is empirical and condition-dependent.

\section{Related Work and Positioning}

LOB forecasting has moved from convolutional and recurrent architectures such as DeepLOB \cite{zhang2019deeplob} toward transformer-style models including ViT-LOB \cite{liu2024vitlob}, LiT \cite{xiao2025lit}, TLOB \cite{tlob2025}, and LOBFrame-style benchmarks \cite{lobframe2024}. These works target predictive modeling and accuracy across horizons, datasets, or regimes. TIP-Search is complementary: it asks how an inference service should select among available predictors when deadline budgets and arrival rates change.

The scheduling side is closer to adaptive inference, model serving, and latency-aware model selection \cite{latencyawareearlyexit2021,teerapittayanon2016branchynet,crankshaw2017clipper,gujarati2020clockwork,romero2021infaas}. Closest systems baselines include RAMSIS and SneakPeek \cite{ramsis2024,sneakpeek2025}, for latency-critical and data-aware model selection. TIP-Search is narrower than full serving systems: it makes a per-request feasible-set decision over trained market predictors and treats RAMSIS/SneakPeek-style evidence as paper-derived systems baselines, not official reproduced external systems. The contribution is a reproducible decision layer over accuracy--latency tradeoffs, not external SOTA model accuracy.

This also separates TIP-Search-OCO-ACPO from pure task-ordering schedulers: EDF and LLF order pending tasks, while TIP-Search chooses heterogeneous predictors using finite-worker state, queue-adjusted budgets, conformal latency margins, online arrival gaps, virtual backlog, deadline debt, shielded experts, and projected constraint duals. CPO and OPT variants are retained for attribution.

\section{Method: Time-Predictable Inference Scheduling}

Let the candidate model set be $M$. Each model $m$ has a profiled latency estimate $L(m)$, an optional conformal latency-quantile buffer $Q_i^{1-\alpha}(m)$, an estimated accuracy $\hat A(i,m)$ for task or profile $i$, and an effective profile sample size $n_{i,m}$ when available. The dispatch configuration also supplies the task-level latency-uncertainty carrier $B^u_i$. The inference service has $K$ workers with availability vector $\mathbf a_i$ when task $i$ arrives at time $t_i$. TIP-Search-OCO-ACPO and its base policies assign the task to the earliest available worker, set $s_i=\max(t_i,\min_k a_{i,k})$, record queue wait $q_i=s_i-t_i$, and define the queue-adjusted budget $B_i=D_i-q_i$, i.e., the remaining budget after queue wait. The router also maintains an online arrival-gap forecast $\bar g_i$ from an exponential moving average of observed gaps, with a configured cold-start default, and a deadline-debt state $Z_i$ that accumulates previous slack deficits. At dispatch time, the deployable source-agnostic router may use

\begin{table}[tb]
\centering
\scriptsize
\setlength{\tabcolsep}{3pt}
\begin{tabular}{p{0.22\columnwidth}p{0.68\columnwidth}}
\toprule
Symbol & Meaning \\
\midrule
$i,t_i$ & Task index and arrival time. \\
$D_i$ & Task deadline budget. \\
$q_i,B_i$ & Queue wait and queue-adjusted remaining budget. \\
$M,m$ & Candidate fixed predictor set and one model. \\
$L(m)$ & Profiled model latency carrier. \\
$B_i^u$ & Task-level latency-uncertainty carrier. \\
$Q_i^{1-\alpha}(m)$ & Conformal latency-quantile buffer. \\
$\hat A^{-}(i,m)$ & Lower-confidence-bound accuracy score. \\
$K,\mathbf a_i$ & Worker count and worker-availability vector. \\
$\bar g_i$ & Online arrival-gap forecast. \\
$W_i$ & Current virtual work in the worker pool. \\
$Z_i$ & Deadline-debt state. \\
$\mathcal E$ & Finite CPO expert set. \\
$\mu_i$ & Projected drop/response/debt dual vector. \\
$g_i$ & Realized constraint vector after dispatch. \\
$\bar c_T^j$ & Mean realized constraint for $j\in\{r,s,d\}$. \\
\bottomrule
\end{tabular}
\caption{Key notation used by the TIP-Search online scheduling policy.}
\label{tab:notation}
\end{table}

\[
\begin{array}{l}
\mathcal I_i=\{D_i,t_i,\mathbf a_i,K,q_i,B_i,\bar g_i,\\
\quad \{L(m),Q_i^{1-\alpha}(m),\hat A(i,m),n_{i,m}\}_{m\in M},B^u_i\},
\end{array}
\]
but not realized future arrivals, realized service times, ground-truth labels, oracle source labels, or post-hoc diagnostic profiles. A deployable online policy maps $\mathcal I_i$ to a model in $M$ or to a drop decision. Source-aware, oracle, and accuracy-profile controls are reported as diagnostics outside this policy class.

The evaluated online objective is timely service under explicit constraint budgets:
\[
\begin{array}{ll}
\max_{\pi\in\Pi_{\mathrm{online}}} & \sum_i {\bf 1}\{\hat y_i=y_i,\tau_i\le D_i\}\\
\mathrm{s.t.} & \bar c_T^r\le b^r,\quad \bar c_T^s\le b^s,\\
& \bar c_T^d\le b^d ,
\end{array}
\]
where $\tau_i$ is realized response time, $g_i^r$ records drop/deadline-miss violation, $g_i^s$ records response or negative-slack pressure, $g_i^d$ records deadline-debt pressure, and $\bar c_T^j=T^{-1}\sum_{i=1}^T g_i^j$ for $j\in\{r,s,d\}$. Labels are used only for evaluation, not dispatch.

Given task $i$, base TIP-Search forms an admissible set from the profiled latency and remaining budget. TIP-Search-OPT and its LD/VQ/Q ablations use a stricter chance-constrained feasible set:
\[
M_i^\alpha = \{m \in M: L(m)+z_\alpha B^u_i+\eta \leq B_i\}.
\]
TIP-Search-CPO uses the conformal latency-quantile feasible set
\[
M_i^{\mathrm{CPO}}=\{m\in M: L(m)+Q_i^{1-\alpha}(m)+\eta\leq B_i\}.
\]
Here $z_\alpha$ is the configured chance-margin multiplier and $\eta$ is the required slack margin. The implemented service sets $Q_i^{1-\alpha}(m)$ from the audited quantile buffer available at dispatch. The OPT chance-margin and CPO conformal filters are audited as separate carriers; they match only under a configured equality between $Q_i^{1-\alpha}(m)$ and $z_\alpha B^u_i$. If the feasible set is empty, the task is dropped. The base TIP-Search rule is recovered when $q_i=0$, $z_\alpha B^u_i=0$, $Q_i^{1-\alpha}(m)=0$, $\eta=0$, and risk penalties are disabled. TIP-Search++ is the one-step risk-aware version used for the executed official timely-service gate. It computes a lower confidence bound accuracy score
\[
\hat A^{-}(i,m)=\hat A(i,m)-\gamma\sqrt{\frac{\log(2/\delta)}{2n_{i,m}}},
\]
with uncertainty weight $\gamma$ and confidence parameter $\delta$. This score is evaluated with $n_{i,m}>0$; when count metadata are missing, the implementation uses a configured positive default effective sample size. The LD layer estimates the finite-horizon load that will share the same worker pool,
\[
\Phi_i(m)=\frac{\max(W_i+L(m)+H L_{\min}-K H \bar g_i,0)}{K B_i},
\]
where $H$ is the virtual lookahead horizon, $W_i=\sum_k\max(a_{i,k}-t_i,0)$ is current virtual work, and $L_{\min}=\min_{m'\in M_i^\alpha}L(m')$ is the fastest admissible service carrier for the anticipated arrivals. After each executed task, the deadline-debt state is updated as
\[
Z_{i+1}=\beta Z_i+\max(\eta-\sigma_i,0),
\]
where $\beta\in[0,1]$ is a geometric decay factor and $\sigma_i$ is realized post-inference slack. This virtual debt update is a finite-horizon drift surrogate: it records repeated slack shortfalls without claiming global queue stability. TIP-Search-OPT adds a bounded projected drop-risk pressure:
\[
\begin{array}{l}
\Delta_i(m)=\frac{\max(W_i+L(m)+H L_{\min}-K\max(B_i-\eta,\epsilon_B),0)}
{K\max(B_i-\eta,\epsilon_B)}.
\end{array}
\]
Here $\Delta_i$ is deterministic projected drop-risk pressure over the worker pool's future slack capacity, with $\epsilon_B>0$ used only after budget exhaustion. The scores below are evaluated only after the nonempty feasible-set check, so their $B_i$ denominators correspond to positive remaining budgets. The nonnegative weights below are fixed dual-style pressure multipliers in the implemented scheduler, not an adaptive dual-update theorem. At dispatch, TIP-Search-OPT selects
\[
\begin{array}{rl}
m^\star=\arg\max_{m\in M_i^\alpha}\; &
\hat A^{-}(i,m)-\lambda\frac{L(m)}{B_i}\\
&-\rho\frac{\max(\eta+B^u_i-(B_i-L(m)),0)}{B_i}\\
&-\kappa\frac{q_i}{B_i}\frac{L(m)}{B_i}\\
&-\nu\frac{W_i}{K B_i}\frac{L(m)}{B_i}\\
&-\omega\Phi_i(m)
-\theta\frac{Z_i}{K B_i}\frac{L(m)}{B_i}\\
&-\zeta\Delta_i(m),
\end{array}
\]
where $\lambda,\rho,\kappa,\nu,\omega,\theta,\zeta$ weight latency, slack, reactive queue, backlog, virtual-backlog, deadline-debt, and drop-risk pressure. TIP-Search-LD sets $\zeta=0$; VQ additionally sets $\theta=0$; Q also sets $\omega=0$; and TIP-Search++ disables queue-pressure terms.

TIP-Search-CPO is the conformal base policy. It uses nonnegative configured dual-style pressure multipliers over a finite receding-horizon service-pressure surrogate. For feasible $m\in M_i^{\mathrm{CPO}}$, it scores
\[
\begin{array}{l}
S_i^{\mathrm{CPO}}(m)=V\hat A^{-}(i,m)\\
\quad-\Lambda^d C^d_i(m)
-\Lambda^r C^r_i(m)
-\Lambda^s C^s_i(m),
\end{array}
\]
where $V$ is the utility weight, $C^d_i$ is normalized deadline-debt pressure, $C^r_i$ is projected drop-risk violation above the configured drop budget, and $C^s_i$ is finite-horizon service/response pressure under the current worker backlog and online arrival-gap forecast. These pressure carriers are bounded before scoring in the implementation. The multipliers $\Lambda^d,\Lambda^r,\Lambda^s$ are predeclared scheduler parameters in the released implementation; pressure still changes online through $C^d_i,C^r_i,C^s_i$, but the paper does not claim adaptive-dual convergence. The implementation uses deterministic bounded scoring and deterministic tie-breaking, so the result is an executable online scheduler rather than an offline oracle. TIP-Search-OPT, LD, VQ, Q, and TIP-Search++ are evaluated as an audited ablation path that removes or changes pressure terms; this is not an exact algebraic reduction claim.

TIP-Search-ACPO is the shielded finite-expert layer. It defines a finite expert set $\mathcal E=\{\mathrm{base},\mathrm{safe},\mathrm{debt},\mathrm{burst},\mathrm{utility}\}$. Each expert is a CPO preset with a different fixed pressure vector, and the `base' expert is exactly the base CPO configuration. Expert $e$ proposes
\[
m_i^e=\arg\max_{m\in M_i^{\mathrm{CPO}}} S_i^e(m).
\]
Here $S_i^e$ has the same CPO score form with expert $e$'s fixed preset weights. Let $r_i(m)=(C^r_i(m),C^s_i(m),C^d_i(m),L(m)+Q_i^{1-\alpha}(m)+\eta)$ be the audited service-risk vector. ACPO accepts a non-base expert only if every component of $r_i(m_i^e)$ is no worse than the base CPO action $m_i^0$ up to configured tolerances; otherwise the action falls back to base CPO. Among accepted experts, ACPO uses the common base-CPO score of the proposed action plus a finite-expert weight term and a bounded debt/backlog activation bonus. After execution, the selected expert weight receives a multiplicative update from realized deadline miss, slack deficit, and deadline-debt loss. The artifact records selected expert, shield status, base CPO model, expert weights, and expert diagnostics in an expert trace. This shielded finite-expert adaptation is an online systems controller, not a regret theorem or adaptive-dual convergence claim.

TIP-Search-OCO-ACPO (online constraint-observing ACPO) adds a conservative projected-dual selector on top of the same shielded expert proposals. It keeps a nonnegative dual vector $\mu_i=(\mu_i^r,\mu_i^s,\mu_i^d)$ for drop, response/slack, and deadline-debt constraints. For each shield-accepted expert proposal, it evaluates the base-CPO utility and subtracts projected constraint cost:
\[
\begin{array}{rcl}
S_i^{\mathrm{OCO}}(e) &=& S_i^{0}(m_i^e)-\mu_i^r C^r_i(m_i^e)\\
&&{}-\mu_i^s C^s_i(m_i^e)-\mu_i^d C^d_i(m_i^e).
\end{array}
\]
The base expert is scored by the same expression, so OCO-ACPO only departs from base CPO when projected constraint cost justifies the action; ties keep the base action. After execution, realized constraint values $g_i=(g_i^r,g_i^s,g_i^d)$ are computed from deadline miss/drop, negative slack, and deadline-debt after the task. The implementation applies the projected update
\[
\mu_{i+1}=[\mu_i+\eta_\mu(g_i-b)]_+
\]
with configured step $\eta_\mu$ and budgets $b$, where $[\cdot]_+$ denotes componentwise positive-part projection. This is an online constraint-observing projected-dual controller with an auditable information set and projected dual trace, not a global regret theorem or adaptive-dual convergence proof.

TIP-Search-SA-OCO-ACPO is the strongly adaptive variant used for the nonstationary stress evidence. It preserves the CPO feasible-set shield and the projected-dual OCO-ACPO cost, but adds an interval stress score $h_i$ computed from queue wait, finite-worker backlog, deadline debt, and arrival-gap compression. Each shield-accepted expert receives a bounded interval bonus $b_e(h_i)$ before deterministic selection. The bonus favors safe/burst/debt experts when stress is high and utility experts when stress is low, while the base expert remains available and the CPO shield still rejects service-risk increases. The implementation records interval id, stress, dynamic-regret proxy, and constraint-violation proxy for every executed or dropped request.

\noindent\textbf{Dispatch rule.}
For each task, TIP-Search-SA-OCO-ACPO assigns the earliest available worker, computes $q_i,B_i,\bar g_i,Z_i,h_i$, forms $M_i^{\mathrm{CPO}}$, drops the task if empty, asks each finite expert for a CPO-style proposal, applies the base CPO shield, scores accepted proposals with projected dual pressure and interval adaptation, and selects the accepted feasible model with deterministic base-preserving tie-breaking. TIP-Search-OCO-ACPO, ACPO, CPO, OPT, LD, VQ, Q, TIP-Search++, and base max-accuracy scheduling are audited ablations.

The rule remains a finite-horizon online surrogate, not a full queueing optimizer: TIP-Search-SA-OCO-ACPO does not imply global queueing optimality, hard real-time behavior, classifier-SOTA upgrades, or universal peak-load guarantees.

\noindent\textbf{Scoped formal properties.}
All formal claims are local to the implemented online information set. The base feasible-set rule is one-step admissible for the stated latency profile, and if the accuracy profile is uniformly within $\epsilon$ of true one-step utility, its selected feasible utility is within $2\epsilon$ of the best feasible model. Recorded binary service metrics use Hoeffding lower bounds as finite-sample evidence. EDF and LLF are task-ordering baselines only; they do not specify heterogeneous model choice.

\begin{table}[t]
\centering
\scriptsize
\setlength{\tabcolsep}{3pt}
\begin{tabular}{@{}r p{0.86\columnwidth}@{}}
\toprule
\multicolumn{2}{@{}l}{\textbf{Algorithm 1: TIP-Search Shielded Online Dispatch}}\\
\multicolumn{2}{@{}p{0.98\columnwidth}@{}}{\textbf{Require:} task $i$, model pool $M$, worker state $\mathbf a_i$, deadline $D_i$, profiles}\\
\multicolumn{2}{@{}p{0.98\columnwidth}@{}}{\textbf{Ensure:} selected model $m_i^\star$ or drop decision}\\
\midrule
1 & \textbf{function} \textsc{Dispatch}$(i,M,\mathbf a_i,D_i)$\\
2 & Assign $i$ to the earliest available worker; compute $q_i$ and $B_i\gets D_i-q_i$.\\
3 & Update $\bar g_i$, $Z_i$, and stress score $h_i$ when SA-OCO-ACPO is enabled.\\
4 & Form feasible set $M_i^{\mathrm{CPO}}$ by the test $L(m)+Q_i^{1-\alpha}(m)+\eta\le B_i$.\\
5 & \textbf{if} $M_i^{\mathrm{CPO}}=\emptyset$ \textbf{then}\\
6 & \quad Record a constraint event and \textbf{return} drop.\\
7 & \textbf{end if}\\
8 & Compute base action $m_i^0\gets\arg\max_{m\in M_i^{\mathrm{CPO}}}S_i^{\mathrm{CPO}}(m)$; set $\mathcal A_i\gets\{m_i^0\}$.\\
9 & \textbf{for all} expert $e\in\mathcal E$ \textbf{do}\\
10 & \quad Compute proposal $m_i^e\gets\arg\max_{m\in M_i^{\mathrm{CPO}}}S_i^e(m)$.\\
11 & \quad \textbf{if} $m_i^e$ passes the componentwise shield against $m_i^0$ \textbf{then}\\
12 & \quad\quad $\mathcal A_i\gets\mathcal A_i\cup\{m_i^e\}$. \hfill $\triangleright$ accept expert\\
13 & \quad \textbf{end if}\\
14 & \textbf{end for}\\
15 & Score each $m\in\mathcal A_i$ by base utility minus projected dual cost; add SA bonus if enabled.\\
16 & $m_i^\star\gets$ deterministic base-preserving tie-break over $\mathcal A_i$.\\
17 & Dispatch $i$ with $m_i^\star$ to the assigned worker.\\
18 & Observe outcome; update worker state, $Z_{i+1}$, expert weights, and $\mu_{i+1}\gets[\mu_i+\eta_\mu(g_i-b)]_+$.\\
19 & \textbf{return} $m_i^\star$.\\
20 & \textbf{end function}\\
\bottomrule
\end{tabular}

\vspace{1pt}
\parbox{0.92\columnwidth}{\scriptsize Online dispatch procedure implemented by the OCO-ACPO/SA-OCO-ACPO controller. Labels and realized service outcomes are observed only after dispatch, and are not used for model selection.}
\end{table}

Queue-aware properties are likewise conditional. Membership in $M_i^\alpha$ gives queue-adjusted latency-margin safety whenever the audited latency uncertainty carrier upper-bounds service time. The finite-horizon virtual backlog term is a virtual backpressure preference: with equal utility/slack terms and $\omega>0$, the lower-backlog feasible model receives the higher VQ score. The deadline-debt state gives a Lyapunov drift-plus-penalty drift surrogate: larger debt increases pressure toward faster feasible models, and without new slack deficits it decays geometrically. The finite-horizon comparator certificate is an ex-post same-trace replay over identical arrivals, deadlines, worker count, and service-time carrier; it reports terminal and prefix gaps for LD versus fixed comparators and is not a regret guarantee.

For CPO, TIP-Search-CPO, TIP-Search-OPT, TIP-Search-LD, TIP-Search-VQ, and TIP-Search-Q share the same finite-worker information set, and the artifact tests an audited ablation path by changing pressure terms and latency carriers. This is not an exact algebraic reduction unless a separate numerical equivalence gate verifies matched settings. If the residuals used for $Q_i^{1-\alpha}(m)$ are exchangeable under the profiled deployment condition, the CPO filter inherits the corresponding conformal marginal latency-coverage statement for the service-time carrier. For ACPO, the shielded finite-expert rule either executes base CPO or a shield-accepted expert action whose audited service-risk vector is componentwise no worse than the base CPO action up to configured tolerances; the expert trace records the selected expert, shield status, base CPO action, and finite-expert weights. For OCO-ACPO and SA-OCO-ACPO, the artifact additionally records projected dual values and selected constraint vectors, and the gates require coverage, changing dual states or interval adaptation, conservative service noninferiority to CPO, nonstationary predictability gain, and scheduling-only claim scope.

For the implemented finite-expert surrogate, the standard Hedge/projected-dual argument gives a scoped certificate rather than a broad queueing theorem. With finite $\mathcal E$, bounded surrogate losses and constraints in $[0,1]$, a fixed shield-accepted feasible action set, and step sizes of order $\sqrt{\log |\mathcal E|/T}$, cumulative surrogate regret to the best fixed shield-accepted expert and cumulative surrogate constraint violation scale as $O(\sqrt{T\log |\mathcal E|})$. With $S$ piecewise-stationary intervals and the recorded interval selector, the dynamic-regret proxy scales as $O(\sqrt{TS\log(|\mathcal E|T)})$ for the same surrogate sequence. These are finite-expert, shield-conditioned statements; they do not imply global queueing optimality, hard real-time schedulability, classifier-SOTA improvement, adaptive-dual convergence, or validity under distribution shift.

\section{Evaluation Methodology}

The primary evidence is the optimized deployable-local rerun and official-SOTA-first h=10 paired suites. The base-pool table uses 150 unique source windows per runtime task stream (50 examples per source) and expands them into 72,000 scheduled observations across four policies; each listed policy therefore has 18,000 repeated dispatch observations over deadline, arrival, and seed conditions. These scheduled observations are not independent market examples. A residual audit uses stride sampling at 200 target windows/source; because FI-2010 supplies 98 valid stride windows under the local loader, each condition has 498 unique source windows, not repeated head slices. It covers 144 conditions and 11,952 tasks per strategy over ACPO/OCO-ACPO/CPO, SneakPeek-style, fastest, and accuracy controls. Condition-level bootstrap confidence intervals are reported; scheduled-task counts are not treated as independent sample size. The class-balanced FI-2010 repair raises base-pool TIP-Search mean accuracy to 0.994 and macro-F1 to 0.996 over its 18,000 scheduled observations. Local-candidate results mirror the same repeated-condition design and remain diagnostics, not external classifier-SOTA evidence.

The external-artifact path is gated separately. Only official TLOB h10 passes the official-SOTA-ready gate; h1/h2/h5 are protocol-blocked; h20/h30/h50/h100 are source-loader-supported but execution-gated; LiT/ViT-LOB lack verified official executable code/checkpoints; LOBCAST lacks local LICENSE/checkpoint/export evidence; and RAMSIS/SneakPeek are paper-derived systems comparators. Local TransLOB evidence-chain coverage is retained only as implementation coverage and is outside the main external-SOTA ranking. BTC predictions have verified label-recomputation provenance, but remain BTC-specific. Only official-executed rows may enter external-SOTA results. The repaired h10 fixed run uses 2,000 official-window tasks; matched official suites compare the calibrated head and TIP-Search++ against fixed h10 over 72 paired conditions and 7,200 tasks per suite. A separate official h10 profiled systems replay uses deterministic latency-profile timing, 72 paired conditions per strategy, 14,400 scheduled observations per strategy, and records its OCO-ACPO pressure profile in provenance.

The SA-OCO-ACPO stress suite is a deterministic systems microbenchmark, not classifier evidence. It replays regime switches, worker-capacity shocks, deadline shocks, latency drift, and accuracy-latency flips over the same task stream and compares CPO, ACPO, OCO-ACPO, and SA-OCO-ACPO under identical arrivals, deadlines, labels, and service carriers. Its gate requires positive timely-service effect size, visible non-base adaptation, nonnegative regret/constraint proxy traces, and scheduling-only claim scope.

Dataset provenance, license/access notes, and run provenance are stored with the evidence package. FI-2010 split handling is required to remain split-safe. Hardware-specific latency profiles must be regenerated for any final deployment claim.

\section{Results}

Table~\ref{tab:strategy} reports the optimized local rerun. We use timely accuracy, $\Pr(\mathrm{correct}\wedge\mathrm{on\ time})$, because market-prediction services benefit only from correct predictions delivered before deadline; deadline-met rate and drop rate are reported as service-feasibility checks. In the base pool, TIP-Search is more accurate than fastest (0.994 vs. 0.331) and random (0.994 vs. 0.552) while keeping comparable deadline satisfaction; its timely accuracy is 0.991. Local-candidate columns show that the same decision layer can use additional executable artifacts, but they are not external classifier-SOTA claims.

Table~\ref{tab:official} separates metric roles: fixed h10 is the classifier baseline, while TIP-Search++ is judged by usable prediction under deadline. Raw accuracy is reported for transparency, not as the router pass criterion. Against fixed official TLOB h10, the validation-fitted calibration head gives a small held-out raw-accuracy gain from 0.505 to 0.510 and matched-suite gain from 0.380 to 0.400 after validation safety checks. Separately, TIP-Search++ sacrifices raw accuracy but improves timely accuracy from 0.156 to 0.239 and deadline-met rate from 0.391 to 0.962; these router results are timely-service evidence, not a classifier-SOTA claim. A profile-corrected router diagnostic gives the calibrated head a validation-derived score, observes 188 calibrated-head selections in TIP-Search++, and still remains raw-accuracy-blocked.

The full-policy suite reaches the same qualitative conclusion: privileged source-aware and oracle diagnostics bound what extra information can buy, while TIP-Search remains stronger than uninformed or latency-only deployable baselines. TLOB exports provide service-time evidence and one repaired FI-2010 h=10 protocol rerun; this is checkpoint execution evidence, not full retraining reproduction or a complete multi-horizon benchmark.

\begin{figure}[tb]
\centering
\includegraphics[width=0.98\columnwidth]{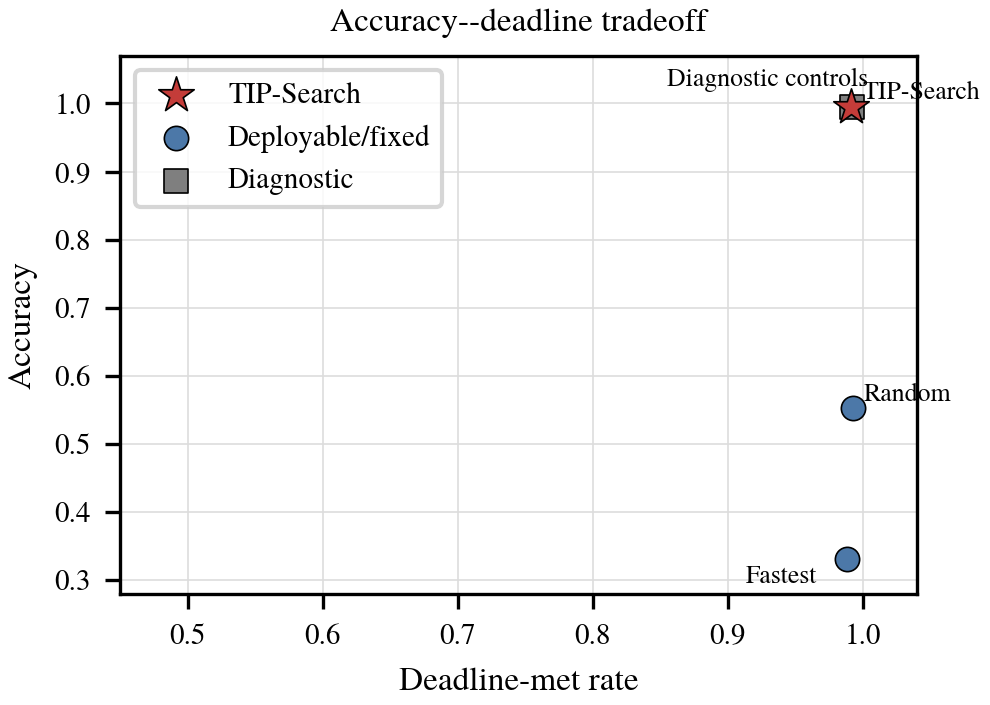}
\caption{Accuracy--deadline tradeoff from the optimized base-pool strategy summary. TIP-Search is shown separately from deployable baselines and diagnostic controls so privileged-information rows are not mistaken for online policies.}
\label{fig:tradeoff}
\end{figure}

Figure~\ref{fig:tradeoff} visualizes the optimized base-pool service tradeoff: TIP-Search preserves deployable deadline filtering while accuracy-profile remains a diagnostic upper control. It is not a claim that unsupported external classifiers were fully reproduced.

\begin{table*}[tb]
\centering
\small
\begin{tabular}{llrrrrrr}
\toprule
Policy & Online & Base Acc. & Base T-Acc. & Base D-Met & Local Acc. & Local T-Acc. & Local D-Met \\
\midrule
Accuracy-profile & no & 0.995 & 0.992 & 0.992 & 0.994 & 0.966 & 0.967 \\
TIP-Search & yes & 0.994 & 0.991 & 0.991 & 0.998 & 0.994 & 0.995 \\
Random & yes & 0.552 & 0.551 & 0.993 & 0.568 & 0.567 & 0.996 \\
Fastest & yes & 0.331 & 0.328 & 0.988 & 0.747 & 0.747 & 1.000 \\
\bottomrule
\end{tabular}
\caption{Optimized submission-scale results over the deployable local subset. T-Acc. denotes timely accuracy, the fraction of tasks that are both correct and on time. Base-pool columns use local DeepLOB checkpoints only; local-candidate columns add executable diagnostic artifacts and are not presented as external classifier-SOTA evidence. Each row aggregates 120 conditions, five seeds, and 18,000 scheduled tasks.}
\label{tab:strategy}
\end{table*}

\begin{table}[tb]
\centering
\scriptsize
\begin{tabular}{lrrrr}
\toprule
Policy & Acc. & T-Acc. & D-Met & Metric role \\
\midrule
Fixed h10 & 0.380 & 0.156 & 0.391 & official fixed model \\
h10 + TIP bias & 0.400 & 0.160 & 0.383 & scoped sanity check \\
TIP-Search++ & 0.244 & 0.239 & 0.962 & router timely service \\
\bottomrule
\end{tabular}
\caption{Official SOTA-first paired comparisons over matched FI-2010 h=10 conditions. The router pass criterion is timely service, not replacement of the fixed classifier's raw-accuracy role. The calibrated head is a validation-fitted wrapper over the unchanged official TLOB h10 checkpoint; TIP-Search++ passes the timely-service gate. Neither result claims h1/h2/h5, execution-gated h20/h30/h50/h100, cross-source, or full external benchmark dominance.}
\label{tab:official}
\end{table}

The comparison should be read by information setting. Accuracy-profile and oracle rows are diagnostics; source-aware dispatch requires reliable source labels; TIP-Search is source-agnostic and deadline-filtered. Component/Pareto artifacts separate local scheduler diagnostics, RAMSIS/SneakPeek-style systems comparators, calibrated-head classifier sanity, and TIP-Search++ timely-service gain. Bootstrap confidence intervals and sign-permutation tests are in the paired artifact. In paired gates, calibrated-head classifier delta is $+0.020$, while the router gate reports $-0.136/+0.084/+0.571$ for raw/timely/deadline deltas. The profile-corrected router diagnostic reports $-0.273/+0.063/+0.555$, so router results remain timely-service evidence. An 864-condition matched h10 profiled systems replay (172,800 scheduled observations) compares fixed h10, fastest, accuracy, utility, risk-aware utility, RAMSIS-style, SneakPeek-style, TIP-Search++, CPO, ACPO, OCO-ACPO, and SA-OCO-ACPO using official h10, calibrated h10 accuracy profiles, and deterministic profiled latency for timing. OCO-ACPO reaches 0.303 timely accuracy and 0.951 deadline satisfaction. Against RAMSIS-style, SneakPeek-style, utility, risk-aware utility, and accuracy controls, paired 72-condition tests show mean deltas of $+0.00285$ timely accuracy (16/4/52 win/loss/tie, sign $p=0.0118$) and $+0.0146$ deadline satisfaction (17/0/55, sign $p=1.5{\times}10^{-5}$); under Poisson/burst pressure the deltas are $+0.00427$ and $+0.0219$. The fastest policy has higher deadline satisfaction (0.989) but much lower timely accuracy (0.187), so it is a latency-only tradeoff rather than a service-quality winner.

The queue-aware audit isolates scheduling behavior from classifier effects using deterministic and live-measured carriers. It covers 720 main-grid settings plus stress/live replay rows. VQ raises timely/deadline rates over Q by $+0.0229$ and lowers wait/p95 response by 0.94/0.63 ms; LD is service-noninferior to VQ and improves deadline-debt stress by $+0.2656$. CPO, ACPO, and OCO-ACPO pass coverage, ablation, trace, conservative-service, and claim-scope gates; traces record 78,617 expert/dual decisions. The LD certificate covers 2,700 matched trace rows with no same-trace failures.

SA-OCO-ACPO is the nonstationary hardening layer. Against CPO, it improves timely/deadline service by $+0.188$, $+0.206$, $+0.267$, $+0.300$, and $+0.417$ under deadline, latency, accuracy-latency, worker, and regime shocks. Non-base adaptation rate is 1.0 in all scenarios, while queue wait, p95 response, drop rate, and deadline debt decrease. The SA gate verifies nonnegative regret/constraint proxy traces and scheduling-only claim scope.

Source-level summaries remain heterogeneous: local-candidate aggregate accuracy is 0.908 for Binance, 0.655 for FI-2010, and 0.917 for LOBSTER-style inputs, with deadline-met rates at or above 0.986. In the stride audit, Binance and LOBSTER-style rows are single-label/label-degenerate service traces; they test latency/scheduling and do not support broad cross-source accuracy claims. The paper therefore reports macro-F1, balanced accuracy, and paired gates rather than a single pooled leaderboard.

\section{Robustness Under Load}

Robustness is checked through finite-sample lower-bound certificates, p95-latency sensitivity, compressed-arrival stress replay, controlled finite-worker load injection, trace-driven queue replay using observed inference times, and live TorchScript service-time measurement. These checks are stricter than mean-latency reporting but remain local evidence rather than live trading deployment.

\begin{center}
\centering
\small
\begin{tabular}{lrrr}
\toprule
Check & Stress & Robust & Metric \\
\midrule
P95 sens. & 1.00$\times$ & 10/12 & 0.906 \\
P95 sens. & 1.50$\times$ & 9/12 & 0.906 \\
Replay & 1.25/0.75$\times$ & 9/12 & 0.750 \\
Replay & 1.50/0.50$\times$ & 4/12 & 0.333 \\
Inject & 2w, 1.00/1.00$\times$ & 10/12 & .167 \\
Inject & 2w, 1.50/0.50$\times$ & 8/12 & .307 \\
Trace & 1w, 1.50/0.50$\times$ & 5/12 & .284 \\
Trace & 2w, 1.50/0.50$\times$ & 9/12 & .027 \\
TLOB service & 4 ckpts & 45/96 & 4.44--4.94 ms \\
\bottomrule
\end{tabular}
\captionof{table}{TIP-Search robustness evidence. The last column is row-specific: P95 rows report deadline-met rate, replay rows report robust rate, injection and trace rows report deadline-miss rate, and TLOB service reports local CUDA service time. Classification evidence is scoped to repaired FI-2010 h=10.}
\label{tab:robustness}
\end{center}

Table~\ref{tab:robustness} makes the overload boundary explicit: nominal and moderate checks are robust, while simultaneous latency inflation and arrival compression remain fragile. These fragile rows define an overload boundary and safe degradation region for condition-dependent systems evidence; they are not hard real-time guarantees. Trace rows show capacity effects, and TLOB horizon 1/2/5/10 TorchScript timing stays timing-only evidence. Robust-router lower bounds above 0.90 strengthen the scoped claim without making it hard real-time.

The queue-aware evidence gate requires controlled load injection, trace replay, live worker timing, online-policy gates, OPT/LD/VQ/Q ablations, projected-dual and expert certificates, nonstationary stress evidence, and a theory wording gate. The hard-strength gate further requires material official h10 timely-service effect, queue/OCO trace depth, SA stress pass, robust lower bounds, Pareto-visible h10 components, the profile-corrected tradeoff diagnostic, and a blocked broad classifier-SOTA boundary. Passing means condition-dependent systems evidence, not global queueing optimality, hard real-time behavior, classifier-SOTA improvement, or live deployment.

\section{Reproducibility and Evidence Chain}

The artifact package includes manifests/provenance, condition/source/strategy/paired/queue-policy/live-worker summaries, SOTA execution gates, external-target audits, claim audit, policy, PDF, readiness checks, and a local reproducibility snapshot with file hashes, git head, and dirty-state enumeration. Key counts include 144,000 optimized rerun tasks, 16,200 TLOB-augmented tasks, 2,000 repaired official h10 tasks, 18,000 profile-corrected h10 diagnostic tasks, 7,200 tasks each in the official router and calibrated-head paired suites, and 172,800 matched h10 profiled systems-replay observations.

The SOTA gate is conservative: only official TLOB h10 is paper-facing FI-2010 classifier evidence. H1/h2/h5 remain protocol-only; h20/h30/h50/h100 remain execution-gated; BTC metric recomputation is scoped to its own prediction arrays; LiT/ViT-LOB, LOBCAST, and SneakPeek remain outside classifier-SOTA claims. The closure gate records zero unsafe claim upgrades.

\section{Limitations}

\noindent\textbf{Boundaries.}
Classifier-facing external evidence is limited to repaired official TLOB h=10 FI-2010. Unsupported TLOB horizons, cross-source TLOB, TransLOB, LiT/ViT-LOB, LOBCAST, RAMSIS, SneakPeek, and BTC broad-use claims remain under the scoped blockers above. Service evidence is historical replay with local checkpoints, load microbenchmarks, live local TorchScript timing, deterministic nonstationary stress replay, and profiling assumptions. The theory is scoped to one-step feasible-set optimality, bounded-error degradation, finite-sample lower-bound evidence, queue-adjusted margins, virtual backlog, deadline debt, finite-horizon certification, conformal/CPO/ACPO/OCO finite-expert statements, and task-ordering separation; it does not establish global queueing optimality, hard real-time guarantees, full multi-horizon TLOB evidence, complete retraining reproduction of modern LOB transformers, or an external SOTA benchmark.

\section{Conclusion}

TIP-Search provides a reproducible, time-predictable inference scheduling study for market prediction under uncertain load. It should be evaluated as a deployment-layer router, not as a new LOB classifier or broad classifier leaderboard. The conservative claim set is supported by the local evidence chain, the OCO-ACPO/SA-OCO-ACPO/ACPO/CPO queue-policy gates, the nonstationary stress suite, the matched h10 profiled systems replay, and the official h10 paired gate, while broad external classifier benchmarking, deployment, and stronger queueing guarantees remain future work.

\bibliography{tip_search_refs}

\begin{thebibliography}{15}
\providecommand{\natexlab}[1]{#1}

\bibitem[{Berti and Kasneci(2025)}]{tlob2025}
Berti, L.; and Kasneci, G. 2025.
\newblock TLOB: A Novel Transformer Model with Dual Attention for Stock Price
  Trend Prediction with Limit Order Book Data.
\newblock arXiv:2502.15757.

\bibitem[{Briola, Bartolucci, and Aste(2024)}]{lobframe2024}
Briola, A.; Bartolucci, S.; and Aste, T. 2024.
\newblock Deep Limit Order Book Forecasting.
\newblock arXiv:2403.09267.

\bibitem[{Crankshaw et~al.(2016)Crankshaw, Wang, Zhou, Franklin, Gonzalez, and
  Stoica}]{crankshaw2017clipper}
Crankshaw, D.; Wang, X.; Zhou, G.; Franklin, M.~J.; Gonzalez, J.~E.; and
  Stoica, I. 2016.
\newblock Clipper: A Low-Latency Online Prediction Serving System.
\newblock arXiv:1612.03079.

\bibitem[{Gujarati et~al.(2020)Gujarati, Karimi, Alzayat, Hao, Kaufmann,
  Vigfusson, and Mace}]{gujarati2020clockwork}
Gujarati, A.; Karimi, R.; Alzayat, S.; Hao, W.; Kaufmann, A.; Vigfusson, Y.;
  and Mace, J. 2020.
\newblock Serving DNNs Like Clockwork: Performance Predictability from the
  Bottom Up.
\newblock In \emph{14th USENIX Symposium on Operating Systems Design and
  Implementation}, 443--462.

\bibitem[{Huang and Polak(2011)}]{huang2011lobster}
Huang, R.; and Polak, T. 2011.
\newblock LOBSTER: Limit Order Book Reconstruction System.
\newblock SSRN.

\bibitem[{Liu and Layland(1973)}]{liu1973scheduling}
Liu, C.~L.; and Layland, J.~W. 1973.
\newblock Scheduling Algorithms for Multiprogramming in a Hard-Real-Time
  Environment.
\newblock \emph{Journal of the ACM}, 20(1): 46--61.

\bibitem[{Liu et~al.(2024)Liu, Sham, Ma, and Fu}]{liu2024vitlob}
Liu, Z.; Sham, C.-W.; Ma, L.; and Fu, C. 2024.
\newblock ViT-LOB: Efficient Vision Transformer for StockPrice Trend Prediction
  Using Limit Order Books.
\newblock In \emph{2024 10th International Conference on Applied System
  Innovation}, 436--438.

\bibitem[{Mendoza, Romero, and Trippel(2024)}]{ramsis2024}
Mendoza, D.; Romero, F.; and Trippel, C. 2024.
\newblock Model Selection for Latency-Critical Inference Serving.
\newblock In \emph{Proceedings of the Nineteenth European Conference on
  Computer Systems}, 874--892.

\bibitem[{Ntakaris et~al.(2018)Ntakaris, Magris, Kanniainen, Gabbouj, and
  Iosifidis}]{ntakaris2018benchmark}
Ntakaris, A.; Magris, M.; Kanniainen, J.; Gabbouj, M.; and Iosifidis, A. 2018.
\newblock Benchmark Dataset for Mid-Price Forecasting of Limit Order Book Data
  with Machine Learning Methods.
\newblock \emph{Journal of Forecasting}, 37: 852--866.

\bibitem[{Romero et~al.(2019)Romero, Li, Yadwadkar, and
  Kozyrakis}]{romero2021infaas}
Romero, F.; Li, Q.; Yadwadkar, N.~J.; and Kozyrakis, C. 2019.
\newblock INFaaS: A Model-Less and Managed Inference Serving System.
\newblock arXiv:1905.13348.

\bibitem[{Tan et~al.(2021)Tan, Li, Wang, Huang, and
  Xu}]{latencyawareearlyexit2021}
Tan, X.; Li, H.; Wang, L.; Huang, X.; and Xu, Z. 2021.
\newblock Empowering Adaptive Early-Exit Inference with Latency Awareness.
\newblock In \emph{Proceedings of the AAAI Conference on Artificial
  Intelligence}.

\bibitem[{Teerapittayanon, McDanel, and
  Kung(2016)}]{teerapittayanon2016branchynet}
Teerapittayanon, S.; McDanel, B.; and Kung, H.~T. 2016.
\newblock BranchyNet: Fast Inference via Early Exiting from Deep Neural
  Networks.
\newblock In \emph{2016 23rd International Conference on Pattern Recognition},
  2464--2469.

\bibitem[{Wolfrath, Frink, and Chandra(2025)}]{sneakpeek2025}
Wolfrath, J.; Frink, D.; and Chandra, A. 2025.
\newblock SneakPeek: Data-Aware Model Selection and Scheduling for Inference
  Serving on the Edge.
\newblock arXiv:2505.06641.

\bibitem[{Xiao et~al.(2025)Xiao, Ventre, Wang, Li, Huan, and Liu}]{xiao2025lit}
Xiao, Y.; Ventre, C.; Wang, Y.; Li, H.; Huan, Y.; and Liu, B. 2025.
\newblock LiT: Limit Order Book Transformer.
\newblock \emph{Frontiers in Artificial Intelligence}, 8: 1616485.

\bibitem[{Zhang, Zohren, and Roberts(2019)}]{zhang2019deeplob}
Zhang, Z.; Zohren, S.; and Roberts, S. 2019.
\newblock DeepLOB: Deep Convolutional Neural Networks for Limit Order Books.
\newblock \emph{IEEE Transactions on Signal Processing}, 67(11): 3001--3012.

\end{thebibliography}

\end{document}